\documentclass{article}
\usepackage[preprint]{colm2026_conference}

\usepackage{microtype}
\usepackage{hyperref}
\usepackage{url}
\usepackage{booktabs}
\usepackage{graphicx}
\usepackage{amssymb}
\usepackage{multirow}

\usepackage{tikz}

\usepackage{lineno}

\definecolor{darkblue}{rgb}{0, 0, 0.5}
\hypersetup{colorlinks=true, citecolor=darkblue, linkcolor=darkblue, urlcolor=darkblue}

\title{A Pilot Study of Autocompleting Tokenizers\thanks{Accepted to the Second Tokenization Workshop (TokShop) at COLM 2026.}}

\author{Samuel Wexler \& Mark Hopkins  \\
Department of Computer Science\\
Williams College\\
Williamstown, MA 01267, USA \\
\texttt{\{saw9,mh24\}@williams.edu} 
}

\newcommand{\circlet}[1]{%
  \tikz[baseline=(char.base)]{
    \node[draw,circle,inner sep=1pt] (char) {#1};
  }%
}

\begin{document}

\ifcolmsubmission
\linenumbers
\fi

\maketitle

\begin{abstract}

Modern input methods routinely rely on autocomplete to omit information that can be recovered from local context. Inspired by these autocomplete-assisted writing systems, we investigate whether Transformer inputs can be compressed in a similar manner. Byte-level tokenization offers a simple and language-independent alternative to subword tokenization, but its longer input sequences typically result in increased computational cost and reduced model quality. We propose a compression scheme that employs a lightweight autoregressive byte language model to identify and remove bytes that are easily predictable from their surrounding context before Transformer processing. The resulting compressed representation is then provided as input to a standard encoder--decoder Transformer. Experiments on machine translation show that a substantial fraction of source-language bytes can be omitted without degrading translation quality. On English--French, our best method preserves translation performance while reducing source sequence length by nearly one-third. Additional experiments on Finnish--English, Russian--English, and Chinese--English demonstrate that the approach generalizes across diverse writing systems and morphological typologies, yielding comparable or improved translation quality at compression ratios between 0.47 and 0.67. These findings suggest that many input bytes are predictable enough to be represented implicitly rather than explicitly, providing a simple mechanism for reducing the sequence-length overhead associated with byte-level models.

\end{abstract}

\section{Introduction}

Subword tokenization \citep{sennrich2016neural} remains the predominant approach for Transformer-based language models \citep{vaswani2017attention,brown2020language,touvron2023llama}, but there has been growing interest in byte-level and tokenizer-free alternatives \citep{xue2022byt5,clark2022canine,tay2022charformer,yu2024megabyte,pagnoni2025blt}. Potential benefits include a smaller embedding matrix (allowing those parameters to be reallocated to the core Transformer layers) and a more uniform treatment of diverse writing systems. However, byte tokenization produces considerably longer token sequences than subword tokenization, which typically results in inferior performance (in terms of runtime, memory consumption, and model quality).

This paper presents a pilot study of a new compression-based input representation that seeks to retain the simplicity and language independence of byte-level representations while mitigating their sequence-length overhead. \textbf{The core idea is to leverage an auxiliary autocomplete model to remove predictable bytes from the original input sequence}.

Humans routinely communicate through autocomplete-assisted input systems that omit information recoverable from context. We investigate whether a similar principle can be applied to Transformer inputs. The approach is specifically inspired by QWERTY keyboard input methods for Chinese \citep{mullaney2024chinese} and the smartphone typing habits of Latin-script language speakers --- both of which typically rely on autocomplete mechanisms. In both cases, the actual sequence of typed characters does not match standard orthography. Instead, it looks like short bursts of Latin letters\footnote{Note that all Latin letters are represented as a single byte with UTF-8.} punctuated by autocompletion triggers. \cite{mullaney2024chinese} refers to this as \emph{hypography} --- from the Greek roots \emph{hypo} (meaning ``below") and \emph{graphy} (meaning ``writing") --- and defines it as ``writing that operates in service of conventional writing or script but at a register beneath." 

\begin{figure}[t]
\begin{center}
\includegraphics[width=\textwidth]{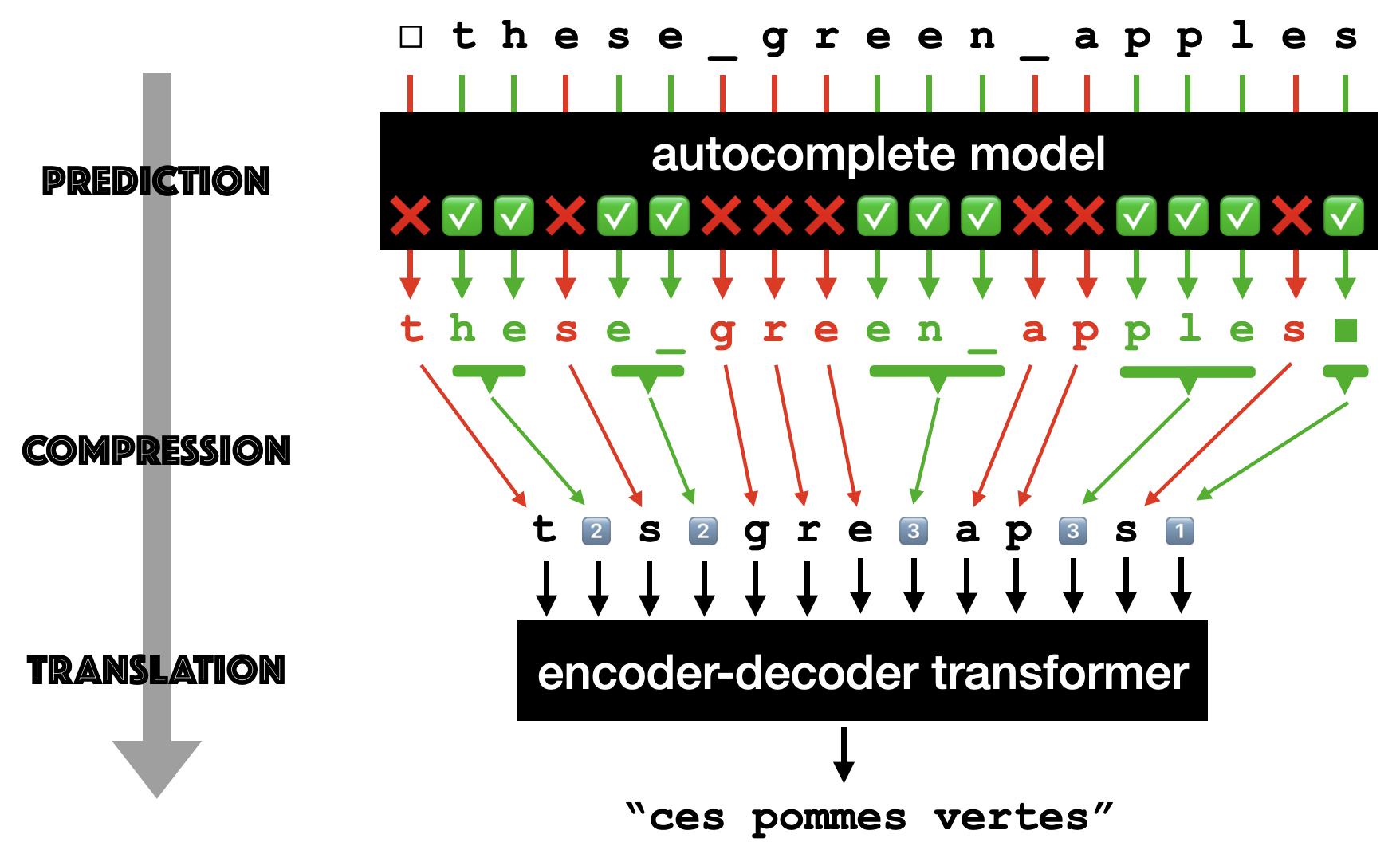}
\end{center}
\caption{One variant (``length-encoded sentinel") of our proposed technique. The original byte sequence is provided as input to an autocomplete model, which auto-regressively predicts the next byte given the previous bytes. Then, byte sequences that were confidently predicted by the autocomplete model are removed. Sentinels are (possibly) inserted to indicate where bytes were removed. Finally, the compressed sequence is ingested by a Transformer.}
\label{fig:demo}
\end{figure}

Figure~\ref{fig:demo} shows one variant of our proposed approach. The general approach has three stages:
\begin{enumerate}
	\item \textbf{Prediction: } The original byte sequence is provided as input to a (small) autocomplete model, which auto-regressively predicts the next byte given the previous bytes.
	\item \textbf{Compression: } Byte sequences that were confidently predicted by the autocomplete model are removed. Sentinels are (possibly) inserted to indicate where bytes were removed.
	\item \textbf{Translation: } The compressed sequence is provided to a (large) Transformer and translated into a desired output.
\end{enumerate}

Our experiments demonstrate that this approach can substantially reduce source sequence length while preserving translation quality. On English--French, we achieve parity with uncompressed byte tokenization after removing nearly one-third of the source bytes. Across additional language pairs spanning agglutinative (Finnish), Cyrillic (Russian), and logographic (Chinese) writing systems, we observe comparable or improved translation performance at compression ratios ranging from 0.47 to 0.67. These findings suggest that a substantial fraction of input bytes are predictable enough to be reconstructed from local context and therefore omitted prior to Transformer processing without degrading translation quality.

Our contributions are as follows:

\begin{itemize}
\item We introduce a compression technique that removes highly predictable bytes prior to Transformer processing.

\item We compare multiple variants of this technique, identifying design choices that best preserve translation quality under aggressive sequence compression.

\item We evaluate the proposed technique across language pairs spanning multiple writing systems and morphological typologies.
\end{itemize}

\section{Related Work}

Byte-level language modeling has emerged as an attractive alternative to subword tokenization. By operating directly on UTF-8 bytes, byte-level models avoid language-specific tokenizers, eliminate out-of-vocabulary issues, and provide a unified representation across writing systems. However, these benefits come at the cost of substantially longer sequences, increasing the computational burden of Transformer architectures.

A substantial body of work has therefore focused on improving the efficiency of byte-level Transformers. ByT5 \citep{xue2022byt5} demonstrated that competitive multilingual performance can be achieved using a pure byte-level representation. Subsequent approaches sought to mitigate the resulting sequence-length explosion through architectural modifications. CANINE \citep{clark2022canine} employs learned downsampling to shorten sequences before deep Transformer processing, while Charformer \citep{tay2022charformer} learns latent segmentations directly from character sequences. MegaByte \citep{yu2024megabyte} and related hierarchical architectures process byte sequences at multiple temporal scales, enabling efficient modeling of long byte streams.

Particularly relevant to our work is Byte Latent Transformer (BLT) \citep{pagnoni2025blt}, which uses a lightweight byte-level model to estimate local prediction difficulty and dynamically partition byte streams into variable-length patches. Regions that are easy to predict are grouped into larger patches, while more surprising regions are represented with finer granularity. Both BLT and our approach exploit the observation that information density varies substantially across a byte sequence and that predictable regions require less explicit representation than unpredictable ones. However, whereas BLT adaptively changes the granularity at which bytes are represented, our method directly removes highly predictable bytes before Transformer processing, yielding a compressed sequence that can be consumed by a standard encoder-decoder architecture without architectural modification.

The work most closely related to ours is the neural compression framework of \cite{lester-etal-2024-compression}. Their approach trains a byte-level language model and uses the resulting probability estimates within an arithmetic coding scheme to generate compressed text representations. To facilitate Transformer training on these compressed streams, they introduce equal-information windows and periodic resetting of both the arithmetic coder and language-model context. Both their approach and our approach exploit the observation that many bytes are highly predictable from local context and therefore need not be represented explicitly. But whereas neural compression seeks information-theoretically efficient encodings that largely discard the original lexical structure, our approach performs selective byte removal while preserving an explicit correspondence between the compressed sequence and the original text.

Our work is also related to approaches that allocate computation according to input difficulty or information content. Prior work has shown that substantial computational savings can be achieved by identifying and removing less informative representations during inference or training. For example, Power-BERT \citep{goyal2020powerbert} progressively eliminates token representations within Transformer layers, while DynamicViT \citep{rao2021dynamicvit} dynamically sparsifies token sequences based on estimated importance. Similar to these approaches, our method seeks to identify portions of the input that can be handled by a lightweight model and reserve the capacity of a larger Transformer for less predictable content. Unlike prior token-pruning methods, however, our approach operates directly on raw byte sequences and performs compression as a preprocessing step, producing a shorter input sequence before any Transformer computation occurs.

Viewed broadly, BLT \citep{pagnoni2025blt}, neurally compressed text \citep{lester-etal-2024-compression}, and our approach all leverage the predictive distribution of a lightweight byte-level model to identify regions of text that require less explicit representation. The three methods differ, however, in how they exploit this signal. BLT preserves the original information content while adapting the granularity of the representation, grouping predictable regions into larger patches and allocating finer representations to less predictable regions. At the opposite extreme, neurally compressed text uses the predictive distribution to construct an information-theoretically compressed encoding that largely replaces the original byte sequence. Our method occupies a middle ground between these approaches: rather than changing the representational granularity or replacing the input with a compressed code, we selectively remove highly predictable bytes while preserving an explicit correspondence between the compressed sequence and the original text. \textit{This perspective places the three approaches along a continuum ranging from adaptive representation, to selective omission, to full entropy-based compression.}

\section{Methodology}

As discussed in the introduction (and visualized in Figure~\ref{fig:demo}), our proposed method first performs autoregressive next-byte \textbf{prediction} on a byte sequence, then \textbf{compresses} the byte sequence based on these predictions, and finally \textbf{translates} the compressed sequence using a standard encoder-decoder Transformer.

\subsection{Prediction}

We train an autoregressive language model $P_\mathsf{ac}(b_k | b_1, \dots, b_{k-1})$ that predicts the $k$th token in a sequence given the previous tokens. In this case, a token is either a byte or a designated start-of-sequence marker $\square$ or end-of-sequence marker $\blacksquare$. In other words, the token vocabulary is $\mathcal{B} = \{0, 1, ..., 255 \} \cup \{\square, \blacksquare \}$. The language model can take many forms, but it should be small and fast -- the main idea is that it will efficiently show us which parts of a document can be predicted without a deep understanding of the text.

In this study, we train a two-layer decoder-only Transformer with $d_\mathsf{model} = 1024$, $d_\mathsf{ff} = 512$ and 16 attention heads per layer.

Then, given input sequence $b_1, \dots, b_m$ (where $b_1 = \square$ and $b_m = \blacksquare$), we compute the following quantities for all $2 \leq i \leq m$:
\begin{eqnarray*}
p_i &=& P_\mathsf{ac}(b_i | b_{<i})\\	
\hat{p}_i &=& \max_{b \not= b_i} P_\mathsf{ac}(b | b_{<i})
\end{eqnarray*}
In other words, we compute the probability $p_i$ of the observed token and the probability $\hat{p}_i$ of the most probable token (apart from the observed token) according to the model. Note that these probabilities can be computed in parallel using a causal mask.

\subsection{Compression}

\begin{figure}[t]
\begin{center}
\includegraphics[width=\textwidth]{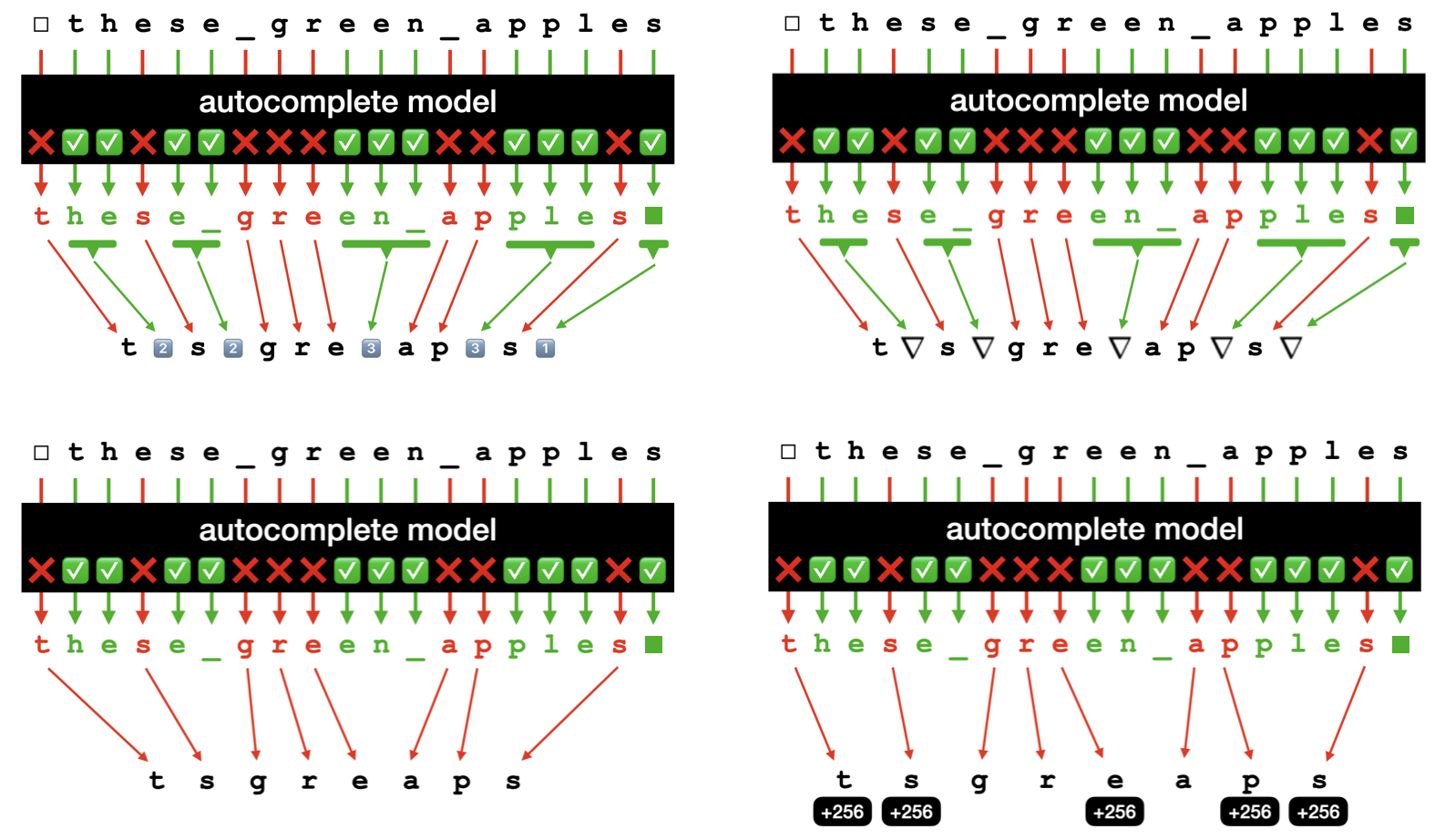}
\end{center}
\caption{Application of the four compression schemes to the segment ``these green apples." From upper left (clockwise): Length-Encoded Sentinel, Explicit Sentinel, Implicit Sentinel, Sentinel-Free.}
\label{fig:compressions}
\end{figure}

Next, we use the probabilities $p_i$ and $\hat{p}_i$ to determine which bytes are ``easily predictable". We experiment with two thresholds:

\begin{enumerate}
	\item \textbf{Absolute Threshold: } Byte $b_i$ is \textit{easily predictable} if $p_i \geq \hat{p}_i$ and $p_i > \alpha$ for hyperparameter $\alpha \in (0, 1]$. The condition $p_i \geq \hat{p}_i$ ensures that the observed byte is the model's most likely prediction.
	\item \textbf{Relative Threshold: } Byte $b_i$ is \textit{easily predictable} if $p_i - \hat{p}_i > \alpha$ for hyperparameter $\alpha \in (0, 1]$.
\end{enumerate}

Once we have determined the indices $\mathcal{E} \subseteq \{2, \dots, m\}$ of easily predictable bytes, we proceed to compress the original byte sequence. We experiment with four compression schemes (examples of each are provided by Figure~\ref{fig:compressions}):

\begin{enumerate}
	\item \textbf{Sentinel-free: } For every $i \in \mathcal{E}$, byte $b_i$ is simply removed from the byte sequence.
	\item \textbf{Explicit Sentinel: } For every consecutive sequence $i, i+1, \dots, j \in \mathcal{E}$, where $i-1 \not\in \mathcal{E}$ and $j+1 \not\in \mathcal{E}$, we replace the byte subsequence $b_i, b_{i+1}, \dots b_{j}$ with a special sentinel $\nabla$.
	\item \textbf{Length-encoded Sentinel: } For every consecutive sequence $i, i+1, \dots, j \in \mathcal{E}$, where $i-1 \not\in \mathcal{E}$ and $j+1 \not\in \mathcal{E}$, we replace the byte subsequence $b_i, b_{i+1}, \dots b_{j}$ with a special sentinel $\circlet{k}$, where $k = j-i+1$ (i.e., the number of bytes replaced).

	\item \textbf{Implicit Sentinel: } For every $i \in \mathcal{E}$, byte $b_i$ is removed from the byte sequence. For every $i \not\in \mathcal{E}$ such that $i+1 \in \mathcal{E}$, we replace $b_i$ with $b_i + 256$.
\end{enumerate}

\subsection{Translation}

We then provide the compressed token sequence $x_1, \dots, x_n$ as input to a standard encoder-decoder Transformer. The input token vocabulary depends on the compression scheme. For the \textbf{sentinel-free} compression scheme, the token vocabulary $\mathcal{V}_\mathsf{sf} = \{0, 1, ..., 255 \} \cup \{\square, \blacksquare, \textsc{Pad} \}$ has 259 tokens: one for each possible byte, plus a start-of-sequence marker, an end-of-sequence marker, and a pad token. The \textbf{explicit sentinel} token vocabulary adds the sentinel $\nabla$ to $\mathcal{V}_\mathsf{sf}$. The \textbf{length-encoded sentinel} token vocabulary adds sentinels $\circlet{1}, \circlet{2}, \circlet{3}, \circlet{4}, \circlet{5}, \circlet{6}, \circlet{7}, \circlet{8}$ to $\mathcal{V}_\mathsf{sf}$ (in this study, we use $\circlet{8}$ to replace any consecutive sequence of easily predictable bytes of length 8 or greater). The \textbf{implicit sentinel} token vocabulary adds $\{256, \dots, 511\}$ to $\mathcal{V}_\mathsf{sf}$.

For this study, we continue to use subword tokenization for the output tokens. Since our experiments focus on machine translation, we use the multilingual tokenizer provided with Meta's NLLB-200 model. Extending the autocompleting tokenizer to the Transformer output is left for future work.

\section{Experiments}

\subsection{Comparing Prediction Thresholds and Compression Schemes}

Our first set of experiments used English-to-French machine translation as a testbed for comparing the effectiveness of different prediction thresholds (absolute and relative) and compression schemes (sentinel-free, explicit sentinel, length-encoded sentinel, and implicit sentinel). We retained 6,994,688 sentence pairs from the WMT14 benchmark \citep{bojar2014findings} whose English source lengths were at most 512 bytes, allowing training without sequence truncation\footnote{All experiments were run on a single Nvidia RTX A6000 GPU with 48GB of RAM.} while training with (uncompressed) byte tokenization. We used the unfiltered WMT14 English-French benchmarks for validation and testing.

\begin{table*}[t]
\centering
\small
\begin{tabular}{c|cc|c|c|c}
\toprule
Compression & \multicolumn{2}{|c|}{Threshold} & Compression & chrF & BLEU \\
Scheme & type & value ($\alpha$) & Ratio & & \\
\midrule

\multirow{6}{*}{Sentinel-Free}
 & Absolute & 0.7 & 0.563 & 57.1 & 29.6 \\
 & Relative & 0.5 & 0.553 & 56.8 & 29.2 \\
 & Relative & 0.3 & 0.500 & 54.0 & 26.5 \\
 & Absolute & 0.5 & 0.486 & 53.1 & 25.9 \\
 & Relative & 0.1 & 0.436 & 48.6 & 21.5 \\
 & Absolute & 0.3 & 0.425 & 47.6 & 20.7 \\
\midrule

\multirow{6}{*}{Explicit}
 & Relative & 0.5 & 0.752 & 59.3 & 32.1 \\
 & Absolute & 0.7 & 0.757 & 58.4 & 31.0 \\
 & Relative & 0.3 & 0.708 & 57.4 & 29.8 \\
 & Absolute & 0.5 & 0.694 & 56.6 & 29.2 \\
 & Relative & 0.1 & 0.647 & 53.7 & 26.1 \\
 & Absolute & 0.3 & 0.642 & 53.8 & 26.2 \\
\midrule

\multirow{6}{*}{Length-Encoded}
 & Relative & 0.5 & 0.752 & 60.1 & 32.7 \\
 & Absolute & 0.7 & 0.757 & 60.1 & 32.8 \\
 & Relative & 0.3 & 0.708 & 59.5 & 32.3 \\
 & Absolute & 0.5 & 0.694 & 59.3 & 31.9 \\
 & Relative & 0.1 & 0.647 & 57.8 & 30.2 \\
 & Absolute & 0.3 & 0.642 & 57.3 & 29.6 \\
\midrule

\multirow{8}{*}{Implicit}
 & Absolute & 0.9 & 0.679 & 60.3 & 33.3 \\
 & Absolute & 0.8 & 0.610 & 59.4 & 32.1 \\
 & Absolute & 0.7 & 0.563 & 59.4 & 32.2 \\
 & Relative & 0.5 & 0.553 & 59.0 & 31.7 \\
 & Absolute & 0.5 & 0.486 & 57.4 & 29.9 \\
 & Relative & 0.3 & 0.500 & 57.0 & 29.7 \\
 & Relative & 0.1 & 0.436 & 55.0 & 27.5 \\
 & Absolute & 0.3 & 0.425 & 54.4 & 26.8 \\

\midrule

\multirow{1}{*}{None}
 & n/a & n/a & 1.0 & 60.4 & 33.2 \\

\bottomrule

\end{tabular}
\caption{Translation quality as a function of compression scheme and threshold. Compression ratio is defined as compressed length divided by original length.}
\label{tab:compression_schemes}
\end{table*}

We trained a standard encoder-decoder Transformer \citep{vaswani2017attention} with 12 encoder layers, 12 decoder layers, $d_\mathsf{model} = 1024$, $d_\mathsf{ff} = 4096$, and 16 attention heads per layer. Models were trained using Adafactor \citep{shazeer2018adafactor} with relative-step learning-rate scaling, weight decay 0.01, and a minibatch size of 32. Training employed automatic mixed-precision arithmetic and gradient-norm clipping with threshold 1.0. Parameters were updated after every minibatch. Validation cross-entropy loss was evaluated every 500 training steps, and the checkpoint with the lowest validation loss was retained. Training was run for 600,000 training steps.

During inference, translations were generated using beam search with a beam width of 4. The maximum output length was set dynamically as $32+3n$, where $n$ is the source sequence length in tokens. Translation quality was evaluated using BLEU \citep{papineni2002bleu} and chrF \citep{popovic2015chrf}. Scores were computed using the Hugging Face \texttt{evaluate} (v0.4.3) implementation of SacreBLEU \citep{post2018call} and chrF, respectively. We report corpus-level BLEU and chrF scores on the test set.

We conducted 26 experiments, each time selecting a different prediction threshold and compression scheme. The results can be found in Table~\ref{tab:compression_schemes}. In that table, the \textbf{compression ratio} is the compressed length divided by the original length of the test set (in bytes). The final row of the table shows the performance of uncompressed byte tokenization.

\begin{figure}[t]
\begin{center}
\includegraphics[width=\textwidth]{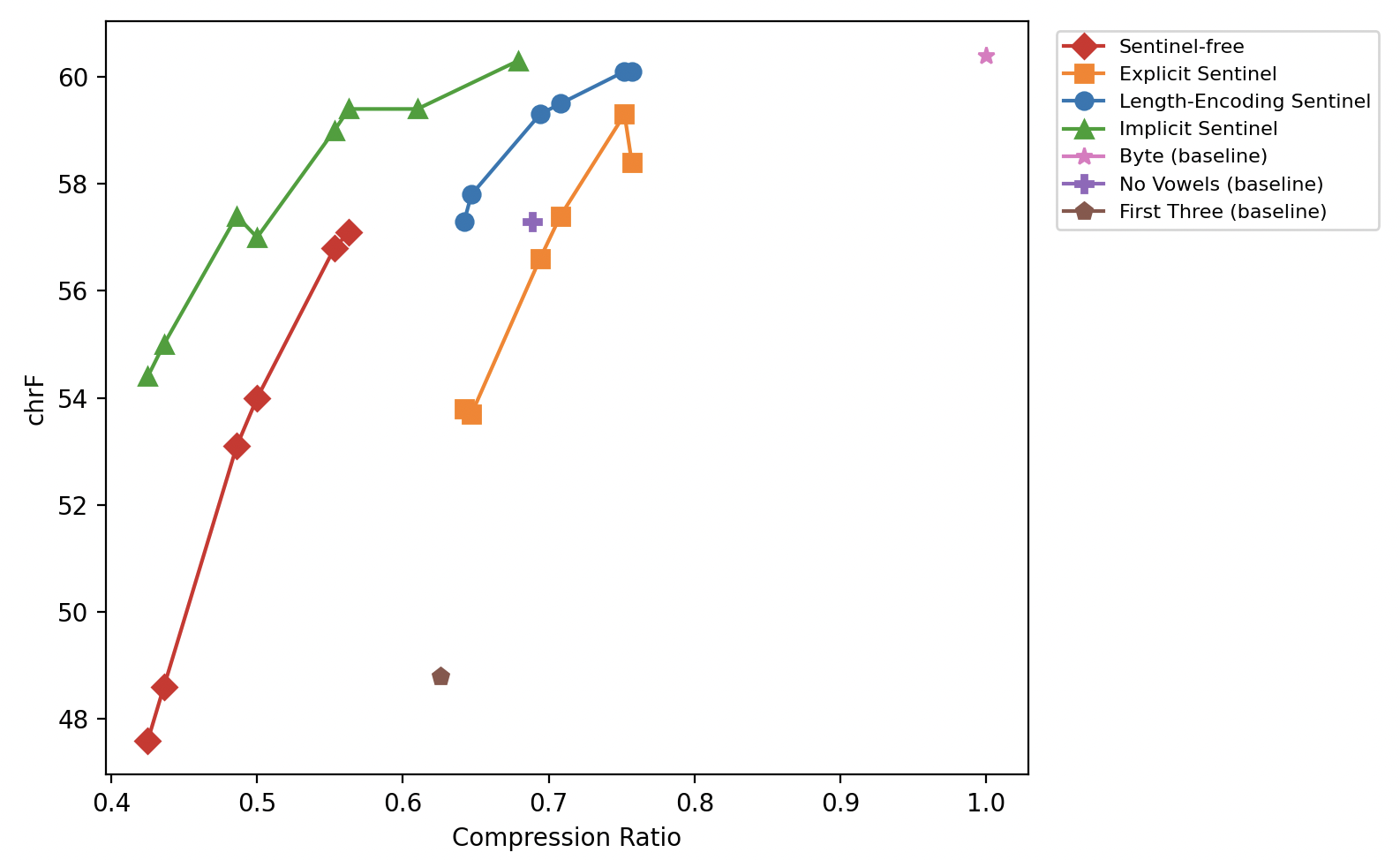}
\end{center}
\caption{Translation quality versus compression ratio for the four compression schemes. Compression ratio is defined as compressed length divided by original length. This plot also includes three baselines: the \textbf{Byte} baseline is uncompressed byte tokenization. The \textbf{No Vowels} baseline removes all vowels from the input sentences. The \textbf{First Three} baseline preserves only the first three letters of each word.}
\label{fig:plot1}
\end{figure}

\begin{figure}[t]
\begin{center}
\includegraphics[width=\textwidth]{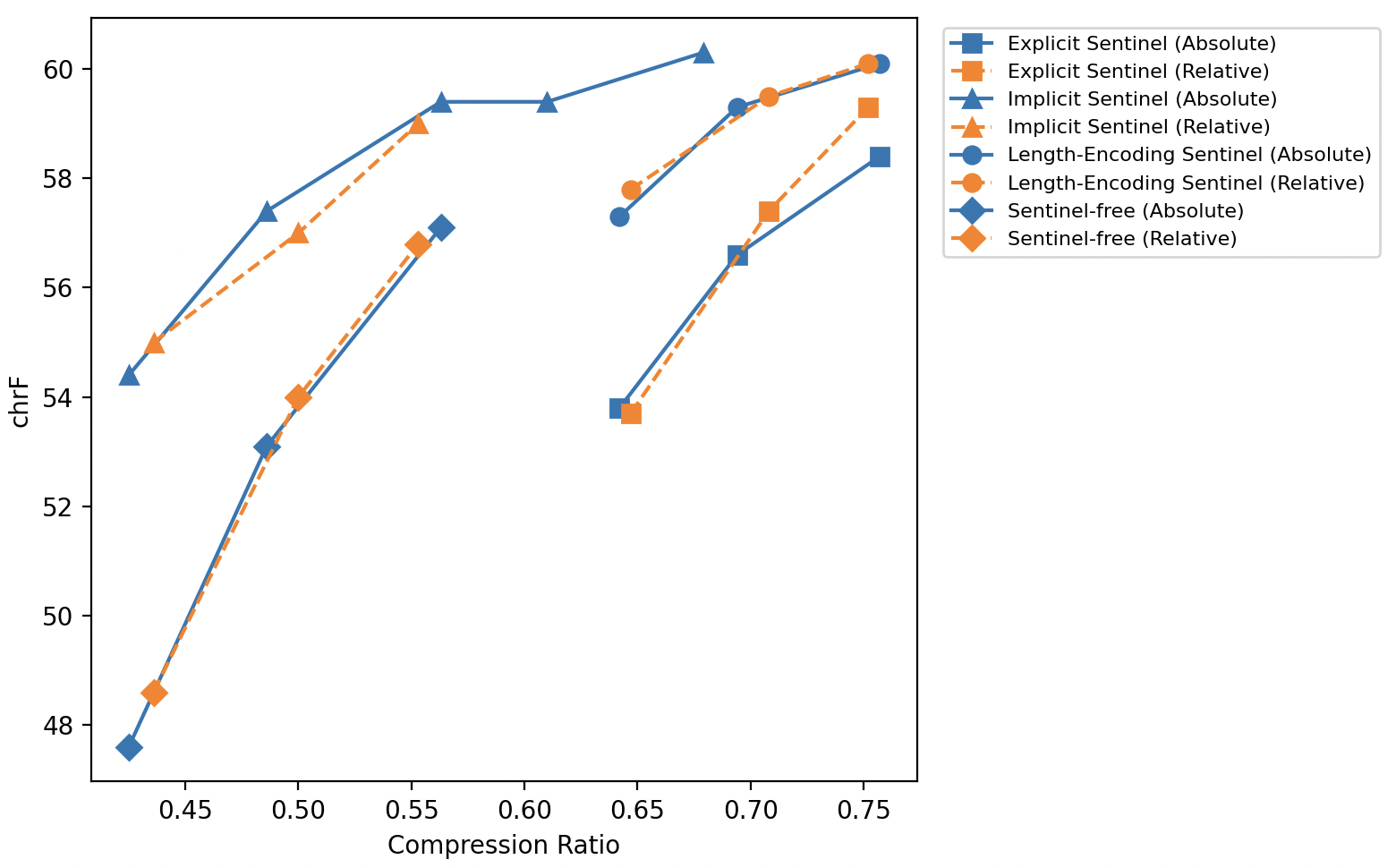}
\end{center}
\caption{Translation quality versus compression ratio for the four compression schemes, with absolute and relative prediction thresholds isolated. There is no clear advantage to absolute or relative prediction thresholds.}
\label{fig:plot2}
\end{figure}

These numbers are visualized in Figure~\ref{fig:plot1} and Figure~\ref{fig:plot2}, which plot each experiment's chrF score versus its compression ratio. Here are some takeaways:

\textbf{The most successful compression scheme is Implicit Sentinel. } Using the Implicit Sentinel compression scheme, we achieve parity with uncompressed byte tokenization at a compression ratio of 0.679, meaning that we can eliminate nearly a third of the English bytes without impacting the quality of the translation system.

\textbf{The autocomplete model is beneficial. } Figure~\ref{fig:plot1} includes two additional baselines besides uncompressed byte tokenization. The \textbf{No Vowels} baseline simply removes all vowels from the input sentences, whereas the \textbf{First Three} baseline preserves only the first three letters of each word. While \textbf{No Vowels} is the more successful of the two, it underperforms the \textbf{Sentinel Free} compression scheme, which removes letters more strategically than these baselines by leveraging the autocomplete model.

\textbf{Encoding the number of removed bytes is beneficial.} The \textbf{Length-Encoded Sentinel} compression scheme outperforms the \textbf{Explicit Sentinel} compression scheme, at the minor cost of adding seven additional sentinels to the token vocabulary.

\textbf{There is no clear advantage to either absolute or relative prediction thresholds.} In Figure~\ref{fig:plot2}, we plot the same data (minus the baselines) but we show which datapoints use absolute prediction thresholds, and which use relative prediction thresholds. They appear to perform similarly.

\subsection{Additional Languages}

Based on the findings from the previous experiments, we applied Implicit Sentinel compression to three WMT19 \citep{barrault2019findings} language pairs: Finnish–English (agglutinative morphology), Russian–English (Cyrillic script), and Chinese–English (logographic script). Because these datasets are larger than the filtered WMT14 English–French subset used in prior experiments, we extended training to 1M steps to ensure convergence under the increased data scale.

In this extended evaluation, some training examples exceed the maximum sequence length supported by the base Transformer architecture, leading to truncation of longer sentences. While this is a standard constraint in training sequence models under fixed computational and memory budgets, it highlights a fundamental limitation of length-bounded processing: when inputs exceed capacity, information must be discarded in a hard and unstructured manner. Our compression-based approach can be viewed as an alternative to this truncation mechanism, in which sequence length is reduced in a data-dependent manner using model-estimated predictability rather than a fixed cutoff. This allows longer inputs to be mapped into the available computational budget in a more adaptive way, while still preserving a correspondence with the original sequence. As such, the setting serves as a useful stress test for comparing hard truncation with learned or heuristic compression strategies under identical resource constraints.

\begin{table*}[t]
\centering
\small
\begin{tabular}{c|c|c|c|c}
\toprule
Language Pair & Compression Scheme & Compression Ratio & chrF & BLEU \\
\midrule
fi-en & Byte & 1.0 &  51.1 &  22.1 \\
fi-en & Implicit Sentinel (abs, $\alpha=0.8$) & 0.646 &  51.2 & 22.1  \\
\midrule
ru-en & Byte & 1.0 &  47.1 & 21.4 \\
ru-en & Implicit Sentinel (abs, $\alpha=0.8$) & 0.468 & \textbf{54.9} & \textbf{28.3} \\
\midrule
zh-en & Byte & 1.0 &  45.3 &  17.3 \\
zh-en & Implicit Sentinel (abs, $\alpha=0.8$) & 0.668 & \textbf{47.9} & \textbf{18.9}  \\
\bottomrule

\end{tabular}
\caption{Translation quality and compression ratio for three WMT19 language pairs. Compression ratio is defined as compressed length divided by original length.}
\label{tab:more_langs_results}
\end{table*}

Experimental results are shown in Table~\ref{tab:more_langs_results}. We observe that Implicit Sentinel compression improves performance on Russian–English and Chinese–English while maintaining parity on Finnish–English. One possible contributing factor is the interaction between compression and script or morphological structure: Russian text in UTF-8 typically requires more bytes per character than Latin-based languages, which may increase the potential benefit of byte-level redundancy removal. Similarly, Chinese text is represented with multi-byte encodings but carries relatively high semantic density at the character level, which may affect the distribution of predictable versus informative regions. Finnish, while morphologically rich, shows similar performance to the byte baseline under the tested compression ratio, suggesting that gains may depend on both script-level encoding properties and language-specific redundancy patterns.

To better understand what information is removed by the autocomplete model, we also analyzed omission rates as a function of byte category. This analysis is deferred to the appendix. 

\section{Conclusion}

We introduced a compression-based approach to byte-level tokenization that leverages a lightweight autoregressive model to identify and remove highly predictable bytes before Transformer processing. Inspired by autocomplete-assisted writing systems, the proposed method shifts some of the burden of sequence modeling from a large translation model to a much smaller byte-level predictor, allowing portions of the input sequence to be represented implicitly rather than explicitly.

Across a range of compression schemes and prediction thresholds, we find that substantial reductions in source sequence length are possible without sacrificing translation quality. In particular, an implicit-sentinel encoding preserves English--French translation performance while removing nearly one-third of the source bytes. Additional experiments on Finnish--English, Russian--English, and Chinese--English demonstrate that the approach generalizes across languages with markedly different writing systems and morphological characteristics, in some cases improving translation quality while simultaneously shortening the input sequence.

Viewed more broadly, our results suggest that byte-level representations contain a considerable amount of predictable information that need not be processed directly by large Transformer models. Rather than treating every byte as equally worthy of computation, future architectures may benefit from allocating computation according to predictability, reserving expensive processing for less predictable portions of the input. In this sense, our work occupies a middle ground between adaptive-representation approaches such as BLT and information-theoretic compression approaches based on arithmetic coding, demonstrating that simple predictability-based omission can be an effective mechanism for reducing sequence length.

Several limitations remain. Our experiments focus exclusively on machine translation and employ a fixed autoregressive predictor architecture. Future work should explore alternative prediction models, applications beyond translation, and more systematic investigations of the relationship between language structure, writing systems, and compression effectiveness. We also leave open the possibility of extending predictability-based compression to both the source and target sides of sequence generation. We hope this pilot study motivates further research into autocomplete-inspired approaches to neural model input.

\bibliography{references}
\bibliographystyle{colm2026_conference}

\appendix
\section{Byte Stream Analysis}

\begin{figure}[t]
\begin{center}
\includegraphics[width=\textwidth]{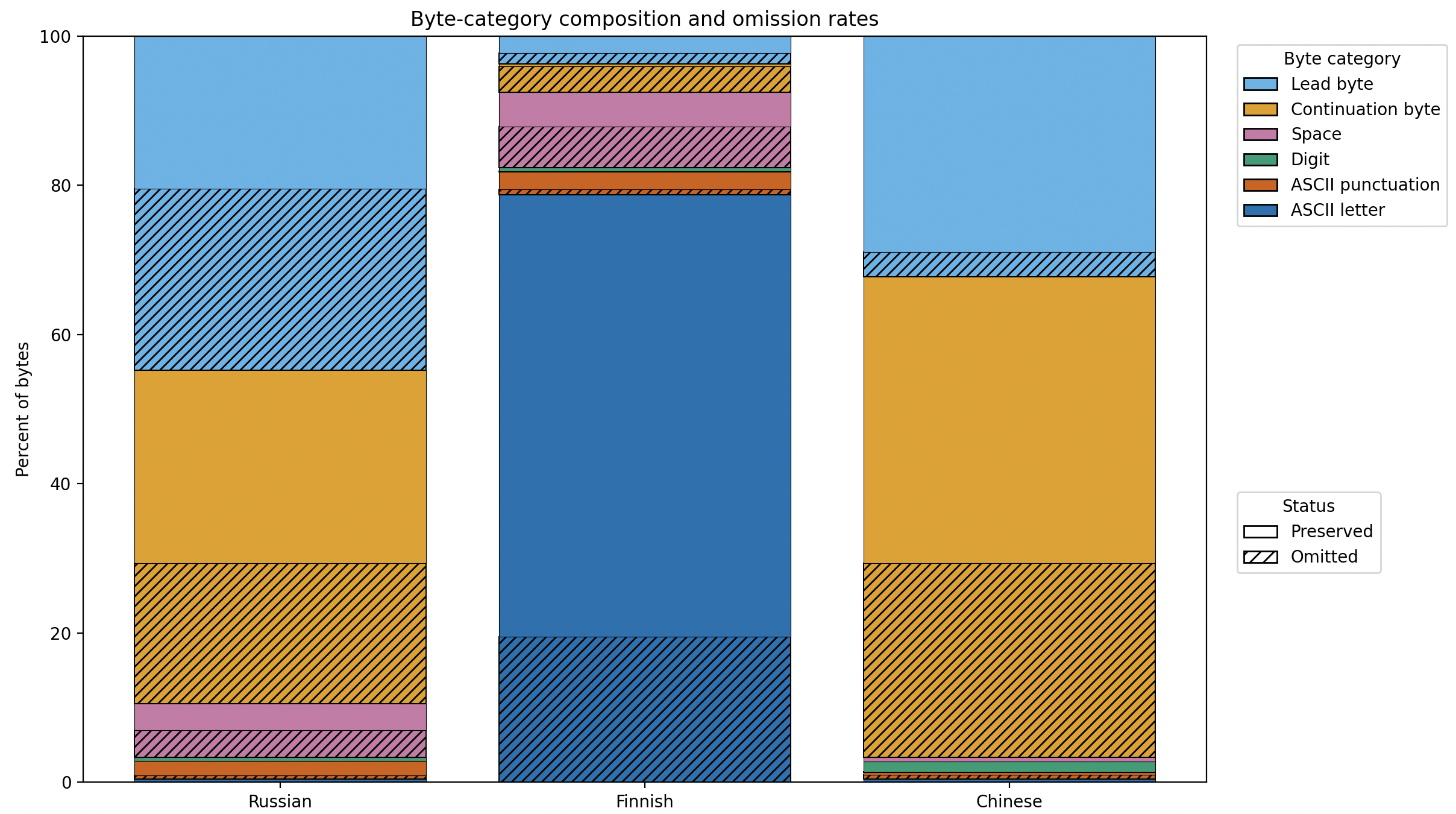}
\end{center}
\caption{Composition of source-language byte streams and omission behavior of the compression step. Each bar represents all source bytes in the test set for a language. Colors indicate byte categories, while hatched regions denote bytes omitted by the autocomplete model. Russian and Chinese contain a substantially larger proportion of UTF-8 lead and continuation bytes than Finnish, and a considerable fraction of these bytes are removed during compression.}
\label{fig:byte_analysis}
\end{figure}

Figure~\ref{fig:byte_analysis} reports the percentage of bytes removed for six categories: ASCII letters, ASCII punctuation, digits, spaces, UTF-8 lead bytes, and UTF-8 continuation bytes.

Several patterns emerge. First, omission rates are consistently highest for structural elements of the byte stream, such as spaces and UTF-8 encoding bytes, rather than for ASCII letters themselves. In Finnish, for example, over 90\% of UTF-8 continuation bytes are removed, while ASCII letters are removed only about 25\% of the time. Similarly, Russian exhibits high omission rates for both UTF-8 lead bytes (54\%) and continuation bytes (42\%), compared to only 12\% for ASCII letters. These results suggest that the autocomplete model preferentially removes bytes that are highly constrained by local context.

Second, the analysis highlights the interaction between our method and UTF-8 encoding. Russian and Chinese text contain large numbers of multi-byte characters, causing UTF-8 lead and continuation bytes to constitute a substantial fraction of the source sequence. Many of these bytes appear highly predictable, likely because they encode orthographic information that is partially determined by neighboring bytes. As a result, compression disproportionately targets encoding-level redundancy in non-Latin scripts.

Third, omission patterns differ substantially across languages. Russian exhibits high omission rates for both UTF-8 lead and continuation bytes, whereas Chinese continuation bytes are removed much more frequently than Chinese lead bytes. Finnish, in contrast, is dominated by ASCII letters and contains relatively few multi-byte characters, limiting the potential gains obtainable through UTF-8 redundancy reduction. These observations are broadly consistent with the experimental results in Table~\ref{tab:more_langs_results}, where Russian benefits most from compression, Chinese shows moderate gains, and Finnish primarily maintains parity with the byte-tokenization baseline.

These findings suggest that the proposed method preferentially removes portions of the byte stream that are highly predictable from local context, including whitespace, punctuation, and components of multi-byte UTF-8 encodings. Understanding the extent to which these effects arise from language structure versus properties of UTF-8 itself remains an interesting direction for future work.

\end{document}